\documentclass[conference,letter]{IEEEtran}

\ifCLASSINFOpdf
\else
   \usepackage[dvips]{graphicx}
\fi
\usepackage{url}
\usepackage{graphicx}
\usepackage{amssymb}
\usepackage{algorithm}
\usepackage{algpseudocode}
\usepackage{amsmath}
\usepackage{booktabs}
\usepackage{array} 

\algdef{SE}[SUBALG]{Indent}{EndIndent}{}{\algorithmicend\ }%
\algtext*{Indent}
\algtext*{EndIndent}

\begin{document}

\title{Adaptive Thresholding Heuristic for \\ KPI Anomaly Detection}

\author{\IEEEauthorblockN{Ebenezer~R.H.P.~Isaac and Akshat Sharma}
\IEEEauthorblockA{Global AI Accelerator, Ericsson,
Chennai, India \\
ebeisaac@ieee.org, akshat.vhs@gmail.com}}


\maketitle

\begin{abstract}
A plethora of outlier detectors have been explored in the time series domain, however, in a business sense, not all outliers are anomalies of interest. Existing anomaly detection solutions are confined to certain outlier detectors limiting their applicability to broader anomaly detection use cases. Network KPIs (Key Performance Indicators) tend to exhibit stochastic behaviour producing statistical outliers, most of which do not adversely affect business operations. Thus, a heuristic is required to capture the business definition of an anomaly for time series KPI.
This article proposes an Adaptive Thresholding Heuristic (ATH) to dynamically adjust the detection threshold based on the local properties of the data distribution and adapt to changes in time series patterns. The heuristic derives the threshold based on the expected periodicity and the observed proportion of anomalies minimizing false positives and addressing concept drift. ATH can be used in conjunction with any underlying seasonality decomposition method and an outlier detector that yields an outlier score. This method has been tested on EON1-Cell-U, a labeled KPI anomaly dataset produced by Ericsson, to validate our hypothesis. Experimental results show that ATH is computationally efficient making it scalable for near real time anomaly detection and flexible with multiple forecasters and outlier detectors.
\end{abstract}

\begin{IEEEkeywords}
Pattern recognition, statistical learning, time series, unsupervised learning, telecom AI. 
\end{IEEEkeywords}

\IEEEpeerreviewmaketitle

\section{Introduction}
\label{sec:introduction}

\IEEEPARstart{I}{nstances} that significantly deviate from the observed statistical pattern in the data are called outliers. Outlier detection is a crucial step in any data science application. Generally, the terms outliers and anomalies are used interchangeably. Nevertheless, within the realm of business, an anomaly can be defined as an unexpected event or situation that pertains to a particular use case. All anomalies may be considered outliers, but not all outliers are anomalies. Network KPI anomaly detection (AD) is an integral part of many use cases in the telecom domain including sleeping cell detection \cite{ming2020ensemble}, ensuring SLA adherence \cite{hong2020machine}, and abnormal traffic detection \cite{alghawli2022complex}. AD has become increasingly important as telecom data continue to grow at an exponential rate. Outlier detectors are usually compared in terms of the area under the receiver operating characteristic curve (AUC-ROC), the greater the area, the better is the tradeoff between true and false positives. However, it is necessary to understand how to set the threshold for a given use case and that this threshold may not be fixed for the life-cycle of the use case. At times, the traffic pattern can change altering the definition of the norm and hence requiring recomputation of the threshold. 

Recently, AD solutions based on Deep Learning (DL) have been gaining popularity \cite{deepNeuralMulti, semiSupStats, chatterjee2022mospat, audibert2020usad}. However, DL methods require extensive computation load on the system. In a typical telecom use case involving cell KPI datasets, the volume of data processed can range from gigabytes to terabytes per month depending on the number of cells which ranges from thousands to tens of thousands. In such cases, employing a DL-based solution would not be viable. It would hence be better to include standard statistical learning or machine learning methods, many of which are described in the PyOD toolbox \cite{han2022adbench}, and then proceed with a use-case-specific filter to mitigate false positives. 

Thresholding in AD defines a limit or boundary to distinguish normal behavior from anomalous behavior in a dataset. A typical K$\sigma$ deviation method involves setting a threshold that is $K$ times the observed standard deviation, $\sigma$, from the mean \cite{kdd2015} based on the assumption that the data is normally distributed. However, this assumption does not hold true for most telecom data. The annual maximum method identifies anomalies based on extreme values observed within a specific time period, usually a year \cite{bezak2014comparison}. It is particularly useful for capturing rare events across longer periods, but not be suitable for detecting anomalies occurring at smaller time scales. LSTM-NDT \cite{lstmNdt} is a deep learning-based method that employs Long-Short-Term-Memory (LSTM) neural networks with non-parametric dynamic thresholding. While LSTM-NDT can effectively capture complex temporal patterns, it may not scale well to large datasets as it is computationally expensive. The Peak Over Threshold (POT) \cite{evt2017} method involves fitting the data to a Generalized Pareto Distribution to determine threshold values. Tuli et al. \cite{tranAD} employed POT and observed better results in comparison to AM. However, POT itself requires a threshold as input to provide a threshold for anomaly detection which appears to be its main disadvantage.\cite{bezak2014comparison}. 


The most successful thresholding methods in literature are computationally intensive, highly sensitive to non-trivial hyperparameter tuning, and few address concept drift. These limitations makes it difficult for existing methods to balance between false positives and false negatives. The method proposed in this article aims to address the above gaps by selecting a threshold that rules out periodic patterns of statistical outliers with an additional mechanism to address concept drift making it robust against noise. The contribution of this article are summarized as follows.
\begin{enumerate}
    \item A time series AD thresholding heuristic specialized for near real time KPI anomaly detection interoperable with any forecaster and any outlier detector that returns a score. 
    \item A business logic to differentiate between outliers and use-case-specific anomalies for telecom KPI datasets.
    \item Introduction of a labelled benchmark dataset for time series AD with telecom KPIs  
\end{enumerate}


\section{Method}
\label{sec:method}

\begin{figure}[t]
\centering
\includegraphics[width=1.0\linewidth]{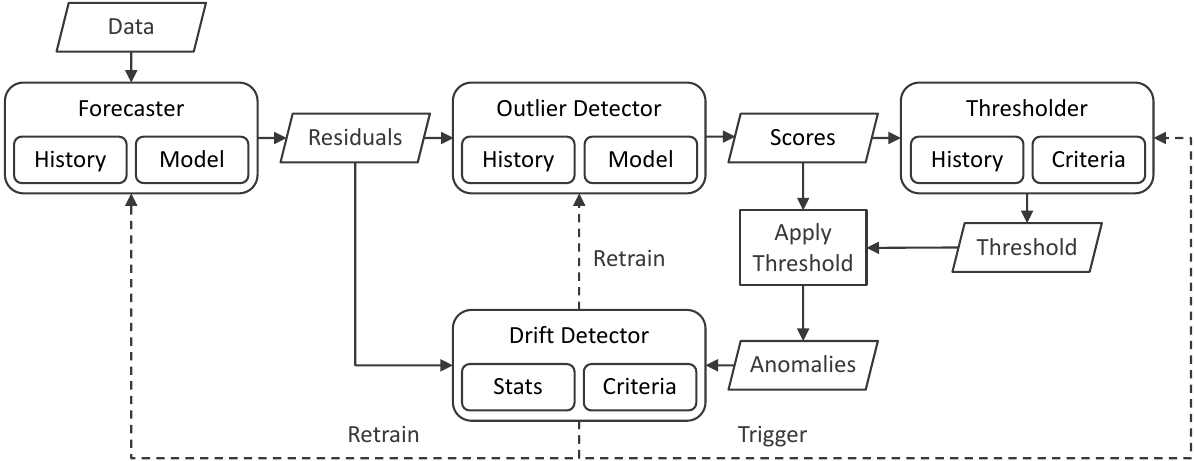}
\caption{Simplified flow of time series anomaly detection with ATH}\label{fig:athflowchart}
\end{figure}

\begin{figure}[t]
\centering
\includegraphics[width=1.0\linewidth]{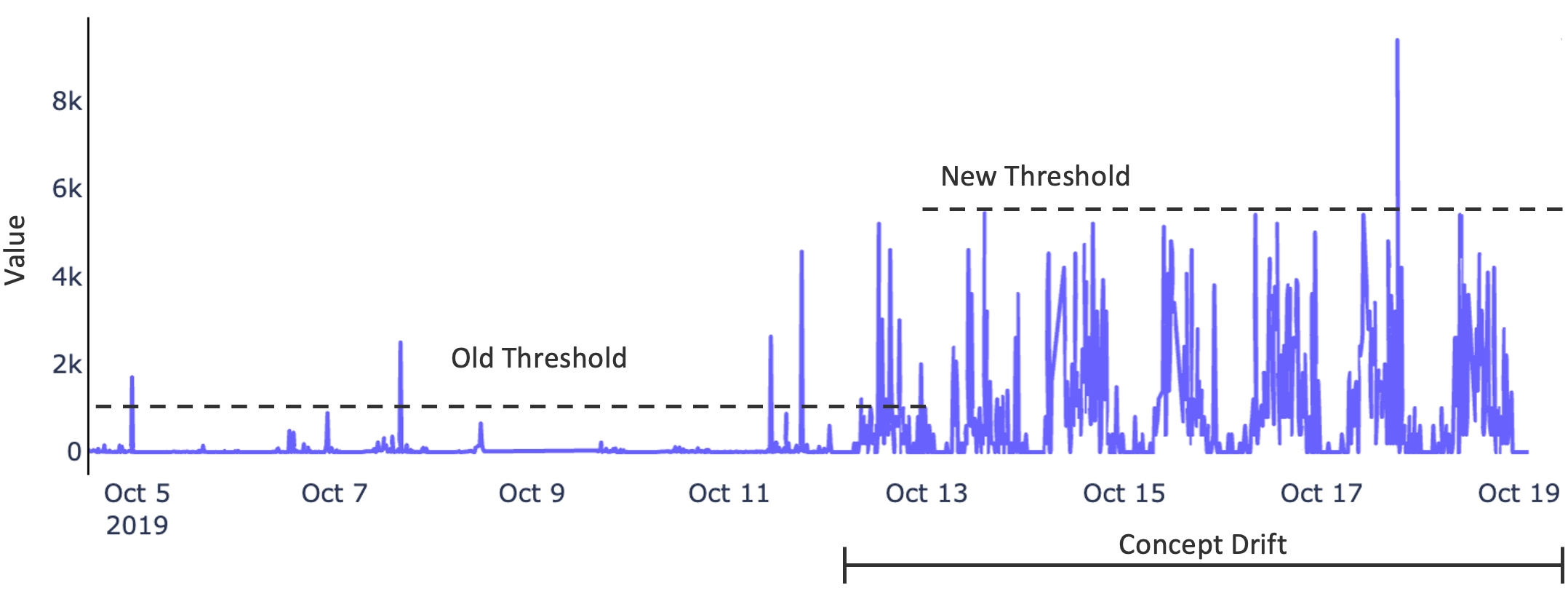}
\caption{Concept drift illustration with ATH. The occurrence of a concept drift breaks the ATH criteria triggering a recomputation of the threshold. }\label{fig:drift}
\end{figure}

This article introduces Adaptive Thresholding Heuristic (ATH), a novel method for setting dynamic thresholds for based on the statistical properties of the data (filed as a patent by Ericsson in \cite{ATHpatent}). It is particularly well-suited to data with unknown or varying distributions. One can assume that anomalous occurrences in time series data does not follow a periodic pattern and are rare in occurrence. In other words, if there is a set of occurrences that seem to be outliers, and if those occurrences occur periodically or too frequently, then the occurrences are most possibly not anomalous. A heuristic is set to adhere to this assumption as constraints. The threshold of a detector is tuned appropriately such that these constraints are satisfied. The occurrence of a concept drift can be identified should the periodicity or anomaly proportion constraint be broken during operation after the threshold is set. Should a concept drift occur, then, depending on the business use case, a simple threshold update can be made by running this routine again, or, in the worst case (as in multivariate systems), the detectors are retrained, and then the heuristic is reapplied.

An overview of the method is shown in Fig.~\ref{fig:athflowchart}. The initial step for time series AD involves the application of forecasting techniques to obtain the residuals. Residuals represent the deviation between the predicted values and the actual observations. These residuals are passed through an outlier detector to obtain anomaly scores. Subsequently, thresholding is employed on these scores, whereby the ATH algorithm is utilized. 
Each stage of the pipeline holds the history of inputs pertaining to a moving window. In a live deployment, the size of the windows may differ from one stage to another. For instance, the forecaster can have an input window of one month while the outlier detector and thresholder can have a window of one week. The drift detector has access to the historical statistics of the percept history of each stage and can trigger a threshold recomputation and retraining should a concept/data drift occur. 




\begin{algorithm}[t]
\scriptsize
\caption{Adaptive Thresholding Heuristic}\label{algo:ath}
\hspace*{\algorithmicindent} \textbf{Input:} \\
\hspace*{3 em} $X$: signal with time series data \\
\hspace*{3 em} $M$: threshold-based outlier detector model \\
\hspace*{3 em} $\mathit{tail}$: either ``left" or ``right" \\
\hspace*{3 em} $\mathit{periodicity\_limit}$: number of permitted periodic outlier occurrences \\
\hspace*{3 em} $\mathit{proportion\_limit}$: permitted proportion of outliers \\
\hspace*{\algorithmicindent} \textbf{Output:} threshold

\begin{algorithmic}[1]
\Procedure{applyATH}{$X, M, \mathit{tail}, \mathit{periodicity\_limit}, \mathit{proportion\_limit}$}
\State Fit the model $M$ using $X$
\State $S \gets$ scores of $X$ using $M$
\State $\mathit{thresh\_list} \gets$ unique values of $S$
\State Sort $\mathit{thresh\_list}$ by its values according to the specified $\mathit{tail}$
\Indent
\State left: ascending order
\State right: descending order
\EndIndent
\State $\mathit{previous\_thresh} \gets$ first item in $\mathit{thresh\_list}$
\For{$\mathit{thresh}$ \textbf{in} $\mathit{thresh\_list}$}
\State $\mathit{outliers} \gets$ empty list
\State Based on the $\mathit{tail}$, do the following
\Indent
\State left: add to $\mathit{outliers}$ all $s \in S$ such that $s <$ $\mathit{thresh}$
\State right: add to $\mathit{outliers}$ all $s \in S$ such that $s >$ $\mathit{thresh}$
\EndIndent
\State $\mathit{diff} \gets$ list temporal differences between each value pair in $\mathit{outliers}$
\State Remove the differences within each consecutive outlier from $\mathit{diff}$
\If{any of following conditions are met} break from the loop
\State Frequency count of any value of $\mathit{diff} > \mathit{periodicity\_limit}$
\State $|\mathit{outliers}|/|X| > \mathit{proportion\_limit}$ 
\EndIf
\State $\mathit{previous\_thresh} \gets \mathit{thresh}$ 
\EndFor
\State $\mathit{final\_thresh} \gets \mathit{previous\_thresh}$
\State \textbf{return} $\mathit{final\_thresh}$
\EndProcedure
\end{algorithmic}
\end{algorithm}

\begin{figure*}[t]
\centering
\includegraphics[width=\linewidth]{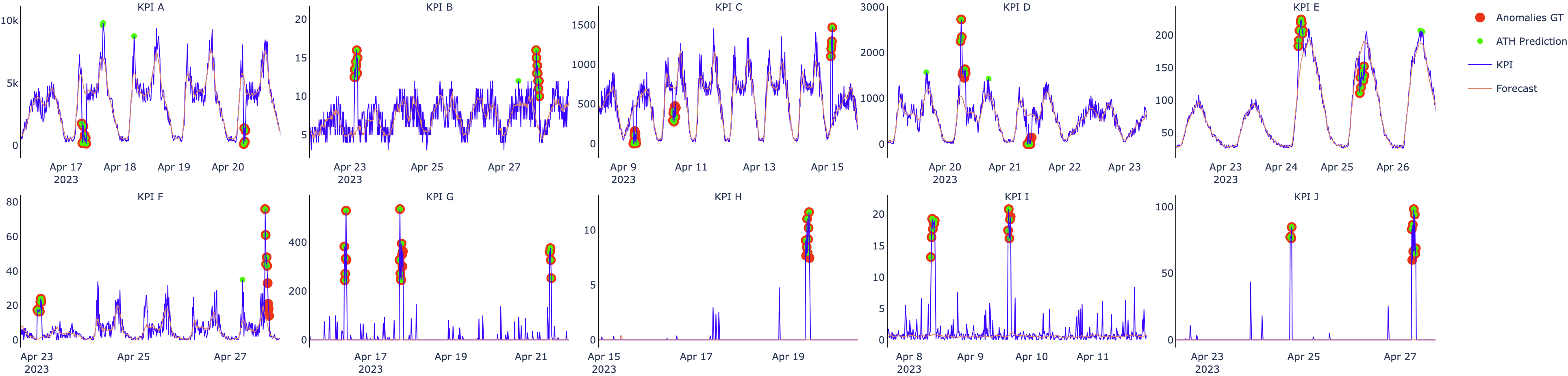}
\caption{Illustration of ATH-based AD through QBSD forecaster and Z-Score detector on the EON1-Cell-U dataset. Only subsets are shown for readability.}
\label{fig:qbsd}
\end{figure*}

A complete description of ATH is provided in Algorithm~\ref{algo:ath}. Given a time series data stream, a detector is fit to a window 
and the outlier scores are extracted. Then, the scores are sorted and checked until the periodicity condition or the anomaly proportion limit is met. Once, the threshold that breaks the constraints is reached, the previous (most recent) threshold in the list that satisfies the constraints is selected as the final threshold. The selected threshold is used to label the anomalies. Both periodicity and proportionality constraints are also monitored through the drift detector so that ATH can be triggered should any of the conditions fail and hence minimize false positives by increasing the threshold accordingly. Also, the drift detector maintains the residual statistics such as its operating range. For instance, reduction in the operating ranges should also trigger ATH to bring down the threshold reduce false negatives. ATH can be applied to any outlier detector that gives a outlier score as output including a simple Z-Score model. 

To check if there is periodicity in the outliers, the system computes the temporal difference between each outlier. This operation can be done in several ways, depending on the use case. For the KPI data, day-wise differences can be considered. For example, if an outlier occurs every day (regardless of the time of day) more than the $\mathit{periodicity\_limit}$, then the observed outlier pattern can be considered periodic. A more fine-grained periodicity check may involve breaking up the given day into multi-hour groups. 

\section{Evaluation}
\label{sec:evaluation}


\subsubsection{Forecasters}

Though ATH does not dictate the forecasting algorithm to be used to obtain the residuals, the effectiveness of time series AD depends on the accuracy of the underlying forecasting method. Quartile Based Seasonality Decomposition (QBSD) (filed as a patent by Ericsson \cite{QBSDpatent}) is a live data forecasting algorithm tailored for telecom data; it does not require an explicit retraining step. 
QBSD has been extensively evaluated in \cite{QBSDpaper} with multiple state-of-the-art forecasters. Since the focus of this paper is not to compare forecasters and detectors but to illustrate the applicability of ATH, only QBSD and Prophet \cite{taylor2018forecasting} are applied for forecasting.

\subsubsection{Outlier Detectors}
The PyOD toolbox \cite{zhao2019pyod} is a Python library of popular outlier detection algorithms. 11 recent detectors in PyOD were evaluated with ATH as the thresholding technique. 6 empirically best performing models were found to be Lightweight On-line Detector of Anomalies (LODA) \cite{pevny2016loda}, Deep One-Class Classification (DeepSVDD) \cite{ruff2018svdd}, Copula-Based Outlier Detection (COPOD) algorithm \cite{li2020copod}, Empirical Cumulative Distribution Functions (ECOD) \cite{li2022ecod}, AutoEncoder \cite{aggarwal2015outlier} and Variational AutoEncoder (VAE) \cite{kingma2013auto}. Only the results of these 6 models will be included in this paper.

The Z-Score is a statistical metric utilized to gauge the deviation of a data point from the mean of its corresponding distribution, expressed in terms of standard deviations. The Z-Score, $Z$, is given by $Z = (x - \mu) / \sigma$, where $x$ is the data point, $\mu$ is the mean of the distribution, and $\sigma$ is the standard deviation of the distribution.

The POT \cite{evt2017} method does not provide a concrete anomaly threshold; it only provides candidate anomalies, i.e., the peaks. ATH can consider each of these peaks as candidate thresholds to provide a better threshold that satisfies the heuristic. The Python implementation of the POT algorithm \cite{potpython} is utilized for this study. 

\subsubsection{Candidate Datasets}
For a univariate anomaly detection problem, the observed variable should possess distinct characteristics that visually differentiates normal and anomalous points when plotted. The definition of an anomaly, according to the use case of interest in this paper, is a rare occurrence that deviates from the typical distribution of the data characterised by an abnormally high peak or abnormally low dip in value. Multiple experiments were conducted utilizing existing datasets such as the AIOps \cite{li2022constructing} and the NAB datasets \cite{lavin2015evaluating}. However, the anomalies labelled in these datasets were found to be inconsistent with the definition of an anomaly as per the scope of this paper. Examples include labelling neighbouring non-anomalous points of an anomalous peak, or labelling a substantial proportion of the input data labelled as anomalies. Consequently, these datasets are not part of the final analysis.

\subsubsection{Selected Dataset}
The dataset utilized for the purpose of the evaluation is referred to as EON1-Cell-U, which is part of the Ericsson Outlier Nexus (EON) \cite{eon1}. EON1-Cell-U is composed of time series KPI designed for univariate AD. There are 10 KPIs in this dataset encompassing different time series characteristics, both seasonal and stochastic. The first 6 KPIs (A through F) exhibit seasonality, that is, a predictable component that is dependent on the time of the day and the day of the week. This pattern can be observed in load KPIs, e.g., Active Uplink Users. The last 4 KPIs (G through J) exhibit stochasticity without seasonality which is characterized by erratic behavior. Such a pattern can be observed by fault monitoring KPIs, e.g., S1 Setup Failure. The KPIs have separate periods for training, validation, and testing; one month each. The interval between two consecutive KPI values is 15 minutes. Training split includes the month of February 2023 (28 days). Similarly, the validation and the test splits include March 2023 (31 days) and April 2023 (30 days) respectively. Anomalies for each KPI have been labeled as either 1, -1, or 0, depending on whether the occurrence is a right-tailed anomaly, a left-tailed anomaly, or not an anomaly respectively. However, these labels are not fed into the algorithm while training or testing (the thresholding is based on unsupervised learning). The labels are used to derive the evaluation metrics from the predictions of the ATH algorithm. 





\begin{table*}
\caption{ATH Performance on EON1-Cell-U in terms of $F_1$ Scores}
\label{tab:athres}
\centering
\begin{tabular}{llccccccccccc}
\toprule
Forecaster & Detector & A & B & C & D & E & F & G & H & I & J & Mean \\
\midrule
Prophet & DeepSVDD & 0.571 & 0.785 & 0.725 & 0.748 & 0.200 & 0.290 & 0.074 & \textbf{1.000} & 0.043 & \textbf{1.000} & 0.544 \\
Prophet & LODA & 0.056 & 0.780 & 0.779 & 0.818 & 0.042 & 0.415 & 0.769 & 0.774 & 0.930 & 0.766 & 0.613 \\
Prophet & ECOD & 0.556 & 0.484 & 0.773 & 0.608 & 0.328 & 0.296 & 0.600 & 0.923 & 0.653 & 0.909 & 0.613 \\
Prophet & COPOD & 0.163 & \textbf{0.984} & 0.424 & 0.250 & 0.273 & 0.436 & 0.889 & 0.923 & 0.952 & 0.837 & 0.613 \\
Prophet & AutoEncoder & 0.467 & 0.608 & 0.697 & 0.696 & 0.328 & 0.471 & 0.741 & 0.750 & 0.930 & 0.766 & 0.645 \\
Prophet & VAE & 0.467 & 0.608 & 0.697 & 0.696 & 0.328 & 0.471 & 0.741 & 0.750 & 0.930 & 0.766 & 0.645 \\
Prophet & Z-Score & 0.485 & 0.933 & 0.767 & 0.727 & 0.319 & 0.391 & \textbf{1.000} & 0.909 & \textbf{1.000} & 0.971 & 0.750 \\
QBSD & COPOD & 0.227 & 0.951 & 0.444 & 0.290 & 0.333 & 0.481 & 0.870 & 0.923 & 0.930 & 0.837 & 0.629 \\
QBSD & ECOD & 0.600 & 0.593 & 0.767 & 0.622 & 0.657 & 0.302 & 0.818 & 0.566 & 0.833 & 0.837 & 0.659 \\
QBSD & DeepSVDD & 0.625 & 0.821 & 0.840 & \textbf{0.870} & \textbf{0.732} & 0.240 & 0.833 & 0.727 & 0.976 & 0.621 & 0.729 \\
QBSD & LODA & 0.625 & 0.941 & 0.757 & \textbf{0.870} & 0.714 & 0.279 & 0.741 & 0.787 & 0.930 & 0.766 & 0.741 \\
QBSD & AutoEncoder & \textbf{0.654} & 0.889 & 0.785 & 0.844 & 0.696 & 0.483 & 0.714 & 0.774 & 0.851 & 0.766 & 0.746 \\
QBSD & VAE & \textbf{0.654} & 0.889 & 0.785 & 0.844 & 0.696 & 0.483 & 0.714 & 0.774 & 0.851 & 0.766 & 0.746 \\
QBSD & Z-Score & 0.636 & 0.901 & \textbf{0.861} & 0.588 & 0.711 & \textbf{0.857} & 0.800 & 0.909 & 0.800 & 0.971 & \textbf{0.803} \\
\bottomrule
\end{tabular}

\vspace{0.7em}
\small The best-performing detector for each dataset has been typeset in boldface for reference.
\end{table*}

\subsubsection{Performance Metric}
$F_1$ score, that is, the harmonic mean of precision and recall, is used as the assessment metric. Though these metrics are usually employed to evaluate supervised learning methods, the availability of the labelled dataset in this problem enables these metrics to be used for this use case. 

\section{Experimental Results and Discussion}
\label{sec:results}

The values for the $\mathit{periodicity\_limit}$ and $\mathit{proportion\_limit}$ were set based on optimal scores on the validation set. The typical value for $\mathit{periodicity\_limit}$ ranged between 2 to 4, while the $\mathit{proportion\_limit}$ ranged between 0.005 to 0.01 (i.e. less than 1\% data to be anomalous). Fig.~\ref{fig:qbsd} depicts the effectiveness of ATH in detecting anomalies with QBSD as the forecaster and Z-Score as the detector with ground truth labels (GT). For seasonal KPIs (A-F), it would be technically possible to apply thresholding directly on the thresholds for the use case concerned. However, stochastic KPIs (G-J) require a deeper focus since these KPIs cannot be forecasted. A standard ``three-sigma-rule" will mark all peaks of a such KPI as outliers even if it is not anomalous. While the periodicity condition addresses seasonal aspects that can be overlooked by forecasters, the proportionality condition addresses the stochasticity ensuring minimal false positives and false negatives.

\begin{figure}[t]
\centering
\includegraphics[width=\linewidth]{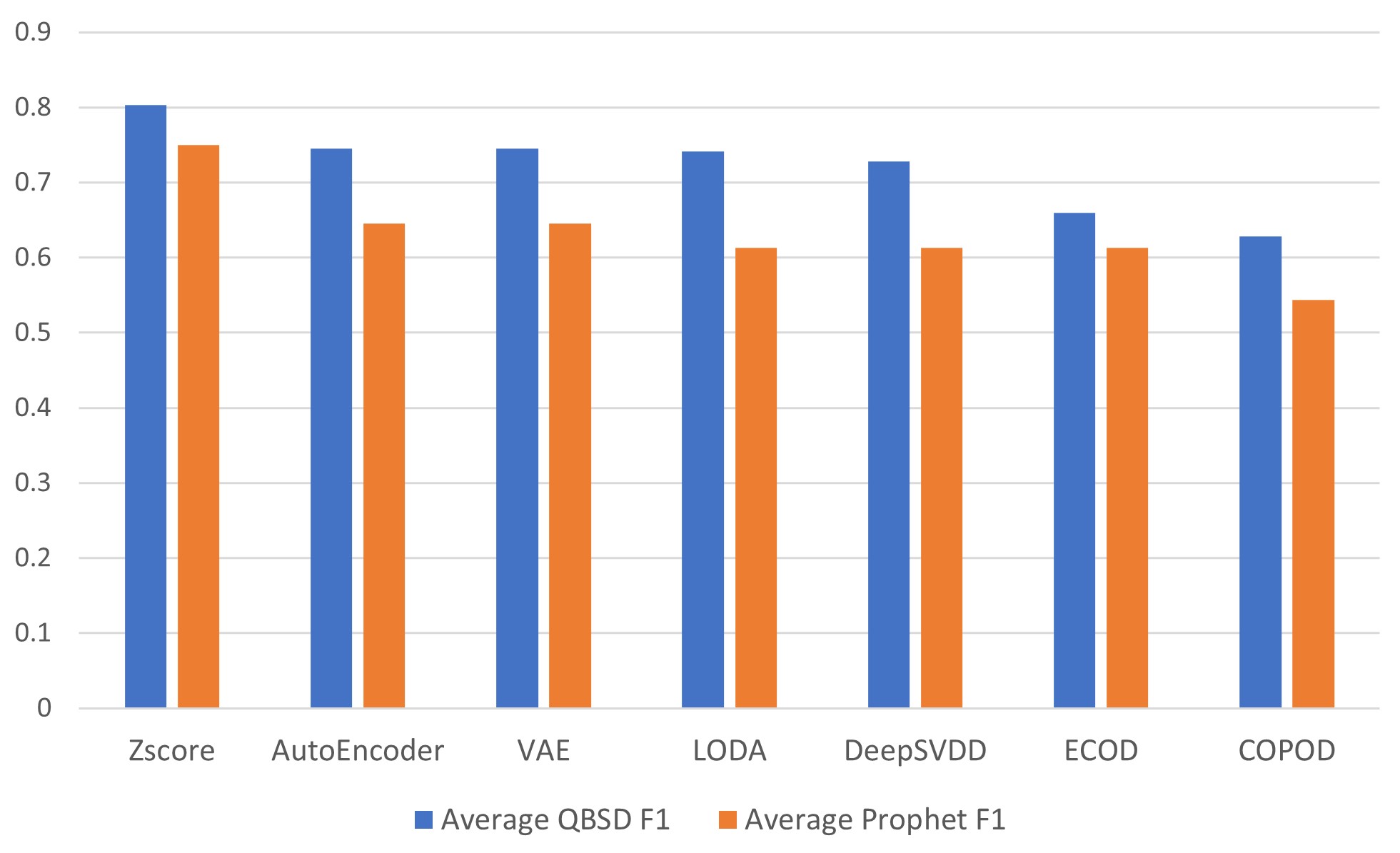}
\caption{Average $F_1$ scores across all KPIs of EON1-Cell-U}
\label{fig:bargraph1}
\end{figure}

The performance metrics computed from the predictions generated by ATH on each forecaster-detector combination for each KPI, were grouped and presented in Table \ref{tab:athres}. The showcased outcomes offer valuable insights into the efficacy of ATH across different forecasters and detectors. DeepSVDD performs remarkably well in some stochastic scenarios (H and J), and some seasonal scenarios (D and E) when coupled with QBSD. Both Autoencoders (AutoEncoder and VAE) gave the same results for all scenarios. 
There is no single forecaster-detector combination can be considered the best for all KPIs. Nevertheless, on average, QBSD forecaster consistently outperforms the results obtained by using Prophet forecaster with the ATH algorithm regardless of the the detector used (Fig.~\ref{fig:bargraph1}). ECOD and COPOD seemed to be the least accurate detectors considered. Surprisingly, the simple Z-Score seems to perform better on average than the other specialized PyOD detectors. This phenomenon can be due to the way the detectors are designed. The outlier score as represented in the Z-Score has a scalar dependency on the magnitude of the anomalous spike. This dependency aligns well with the definition of an anomaly in the dataset considered and the working of ATH. The outlier scores produced by the PyOD algorithms considered in this experimentation does not always follow this linear dependency.

An effective AD solution should also be efficient to make it viable for business, especially in the context of big data applications. To derive the computational complexity of ATH, let $n$ be the number of elements in the input window. Sorting is done in $O(n)$ time using Quicksort. Since the anomaly proportion limit is obviously far less than $n/2$, only $O(\text{log}~n)$ scores are candidates for thresholds. Each of the $n$ elements are processed for every threshold candidate. Temporal difference between each outlier is computed in quadratic time to account for the periodicity check. Thus the overall computational complexity of ATH can be considered to be $O(n\log^2 n)$.

\section{Conclusion}
\label{sec:conclusion}

The proposed ATH algorithm offers a computationally efficient solution to time series AD. It incorporates business logic to differentiate between outliers and use-case-specific anomalies. Its key advantage is its ability to accommodate different outlier detectors and residual extraction methods, making it compatible with a wide range of AD algorithms. The experimental results prove its practical significance in telecom KPIs. The next step would be to extend this heuristic to capture more complex KPI interactions for special cases of anomalies. 


\bibliographystyle{IEEEtran} 
\bibliography{ATH}

\end{document}